# Relational Reinforcement Learning in Infinite Mario

## Shiwali Mohan and John E. Laird

Computer Science and Engineering, University of Michigan
Ann Arbor, MI 48109-2121
{shiwali, laird}@umich.edu

**Abstract**

Relational representations in reinforcement learning allow for the use of structural information like the presence of objects and relationships between them in the description of value functions. Through this paper, we show that such representations allow for the inclusion of background knowledge that qualitatively describes a state and can be used to design agents that demonstrate learning behavior in domains with large state and actions spaces such as computer games.

## Introduction

Computer Games have continuous, enormous state spaces, large action spaces and are characterized by complex relationships between their components. Without applying abstractions, learning in a computer game domain becomes infeasible. Through this work, we investigate some designs that facilitate tractable reinforcement learning in symbolic agents operating in complex domains. We show that imposing hierarchies on the actions and the tasks constricts the state space as a result of which, learning is faster. We further demonstrate that a relational representation allows the use of structural formation such as existence of objects with certain properties or relations between objects in the description of the derived policy.

Soar (Laird, 2008) is a symbolic architecture that has been used to design intelligent agents in many computer game domains. Reinforcement learning has been recently implemented in the architecture and we are exploring agent designs that enable a symbolic agent to learn state-action associations by exploring a new environment while optimizing the expected reward. We have been working with Soar-RL (Nason and Laird, 2005) agents operating in Infinite Mario domain from RL Competition 2009.

## State Representation

As input from the environment, the agent has access to a 16x22 char array each element of which corresponds to a tile on the visual scene and the value of the element corresponds to the type of the tile (coin, brick etc.). The agent also has access to the locations of different types of monsters (including Mario) on the scene, along with their vertical and horizontal velocities. The reward structure varies across different instances of the game.

As a measure of providing structure to the knowledge a Soar agent has about the environment, we moved from the low level tile by tile representation of the visual space to an encoding that is composed of objects and their relationships with Mario and with each other. The knowledge that 'the tile at position (x,y) is of type (t)' is converted to 'there exists a pit at a distance of three tiles in horizontal direction' via elaboration rules in Soar. A similar representation was implemented by Diuk et al. (2008) to solve a real-life videogame, *Pitfall*.

The MarioSoar[1] agent associates unique symbols to different objects in the environment and uses them to reason about and plan its way through a trial. This symbolic representation also allows for addition of more facts and background knowledge in the agent. Rules like 'if there is a pit ahead, then jump while moving right' can be easily encoded. An Infinite Mario episode contains many objects - monsters, coins, question blocks, pipes, raised platforms, and the finish line. Once these objects have been identified from the visual scene, attributes that describe their positions relative to Mario are added to the representation.

## Hierarchical Reinforcement Learning

We have implemented a hierarchy of operators based on a GOMS analysis of Mario (John et al., 1990). The authors demonstrated that the behavior of a Soar agent that used simple, hand-coded heuristics was predictive of the behavior of a human expert playing the game of Mario. The authors distinguish between *key-stroke level operators* (KLO) and *functional level operators* (FLO). A KLO is a primitive, atomic action available to a human user; an action that can be performed using a keyboard or a joystick. In Infinite Mario, these actions include moving right or left, jumping, and shooting. As shown in Figure 1, a FLO is a collection of KLOs that when performed in succession, perform a specific task related to an object, such as killing a monster, or grabbing a coin.

---

[1] http://www-personal.umich.edu/~shiwali/MarioSoar/MarioSoar.html

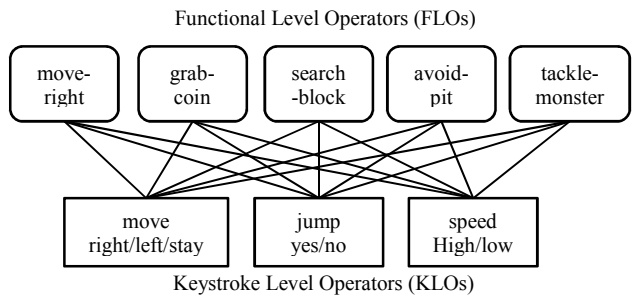

Figure 1: Operator Hierarchy

The GOMS analysis makes two important observations: to successfully complete a game only a small set of FLOs are necessary, all of which can be readily identified; and human experts use local conditions to select between the applicable FLOs. In our implementation, we converted distance between Mario and other entities on the screen to attributes like *isthreat* (for monsters), *isreachable* (for coins and blocks) through elaboration rules. This qualitative information is then used by the agent to make a selection from the group of proposed FLOs.

Soar operators are proposed whenever they can be legally applied. For example, tackle-monster FLO is proposed when a monster is close by and the attribute *isthreat* is *yes*. A FLO proposal causes creation of a substate, in which KLOs are proposed and the agent performs a series of atomic actions. The agent moves out of a substate when the object that caused it is no longer present, which in this case is when Mario kills the monster. Soar RL rules generate numeric preferences for operators. In case of a tie, where multiple objects cause proposals of different operators, the agent uses the numerical preferences associated with the operators to break the tie; the operator with the largest numerical preference is selected and applied.

Our implementation differs from the GOMS analysis in that the agent learns these preferences through its reinforcement learning mechanism whereas the preferences in GOMS analysis were hand-coded by the programmer based on the instruction booklet of the game. The numerical preference is closely linked to the value function for a given state and the proposed operator. We have used SARSA (Rummery and Niranjan, 1994) for Soar agents for this domain.

The agent learns at two distinct levels; at a higher level, the agent learns to select between FLOs when multiple choices present themselves. Once a FLO has been selected to be employed, the agent also learns to select between the correct KLOs to successfully complete the current subtask.

This formulation divides the game into several subtasks of collecting a coin, killing a monster etc. FLOs proposed in a given state allow the agent to interact with corresponding objects in the environment. The agent selects between the FLOs based on which interaction is most immediate. For example if a coin and a monster are in close vicinity of Mario, the agent chooses to deal with the nearby monster first, because a failure to do so would result in the termination of the game. Note, that this knowledge is learned over successive trials of the game.

## Results

Figure 2 shows the Soar-RL agent's performance as it learns selection knowledge for both the FLO's and the KLO's, in comparison to a hand-coded sample agent that was distributed with the Mario environment.

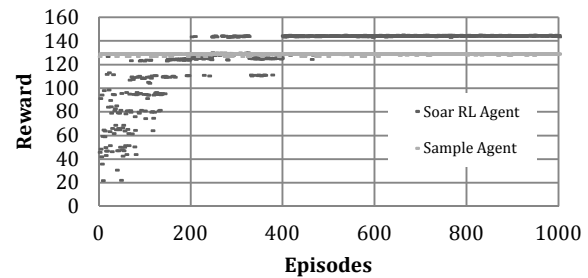

Figure 2: Performance of Soar Rl agent and Sample Agent in level type 0, difficulty 0

## Conclusions and Future Work

The current agent design assumes that the agent can interact with different objects in its vicinity independently. This assumption holds in scenarios where interacting with one object holds more importance than dealing with other near-by objects, e.g. when a coin and a monster are in vicinity, dealing with the monster holds precedence over grabbing the coin. However when, multiple objects have similar importance, e.g. when two monsters are nearby; the assumption breaks down and the agent fails to learn any useful strategy.

We are currently looking at the problem in addition to learning operator preferences, the agent learns new operators and rules as novel situations present themselves.